\def\year{2022}\relax
\documentclass[letterpaper]{article} 
\usepackage{aaai22}  
\usepackage{times}  
\usepackage{helvet}  
\usepackage{courier}  
\usepackage[hyphens]{url}  
\usepackage{graphicx} 
\usepackage{verbatim}
\usepackage{natbib}  
\usepackage{caption} 
\DeclareCaptionStyle{ruled}{labelfont=normalfont,labelsep=colon,strut=off} 
\usepackage{algorithm}
\usepackage{algorithmic}

\usepackage{newfloat}
\usepackage{listings}
\floatstyle{ruled}
\newfloat{listing}{tb}{lst}{}
\floatname{listing}{Listing}
\usepackage{kotex}
\usepackage{subcaption}
\usepackage{multirow}

\title{FLHub: a Federated Learning model sharing service}
\author{
    Hyunsu Mun, Youngseok Lee
}
\affiliations{
    Chungnam National University, Daejeon, Republic of Korea\\
    \{munhyunsu, lee\}@cnu.ac.kr
}

\begin{document}

\maketitle

\begin{abstract}
As easy-to-use deep learning libraries such as \texttt{Tensorflow} and \texttt{Pytorch} are popular, it has become convenient to develop machine learning models.
Due to privacy issues with centralized machine learning, recently, federated learning in the distributed computing framework is attracting attention.
The central server does not collect sensitive and personal data from clients in federated learning, but it only aggregates the model parameters.
Though federated learning helps protect privacy, it is difficult for machine learning developers to share the models that they could utilize for different-domain applications.
In this paper, we propose a federated learning model sharing service named Federated Learning Hub (FLHub).
Users can upload, download, and contribute the model developed by other developers similarly to GitHub.
We demonstrate that a forked model can finish training faster than the existing model and that learning progressed more quickly for each federated round.
\end{abstract}

\section{Introduction}

As deep learning libraries such as \texttt{TensorFlow}, \texttt{Keras}, and \texttt{PyTorch} are popular, it has become easy for non-experts to develop machine learning models and their applications.
In addition, transfer learning is also helpful to develop a high-accuracy model with a small amount of data from a pre-trained model with public data such as \texttt{ImageNet} \cite{pan2009survey}.
Due to public data such as \texttt{MNIST} and \texttt{Tensorflow Datasets}, and public models such as \texttt{ResNet} and \texttt{Inception}, machine learning service developers can easily make their applications.
However, it is difficult for each developer or institution to build their training models for their customized service from scratch. 

Many researchers and companies presented A distributed machine learning framework called federated learning for privacy-sensitive applications such as health care, home IoT services, and personal behavior prediction \cite{li2021ditto}.
Each client keeps data on the device while communicating with the server only for learning parameters in federated learning.
With Google's keyboard application, \texttt{GBoard}, a federated learning client, learns the user's word usage patterns to predict the next keyword \cite{yang2018applied}.
\texttt{IBM}, which acquired Merge Healthcare, aims to predict disease by learning medical and family history, clinical research, and trials and outcomes of hospital patients without exposing personal information \cite{rieke2020future}.

As federated learning is essential for institutions dealing with privacy such as medical, driving, voice, and facial data, it is difficult for each institution to share data.
Though training data and models are publicly available with open-source machine learning libraries in Tensorflow or Keras, it is not easy to share training models in federated learning.
While each FL client has different amounts of data, computing power, and learning schedule, the client performs the learning process in federated learning asynchronously \cite{chen2020asynchronous}.
In addition, there is no convenient tool for developers to exchange federated learning machine learning models.

\begin{figure}[htb]
	\centering
    \includegraphics[width=0.99\columnwidth]{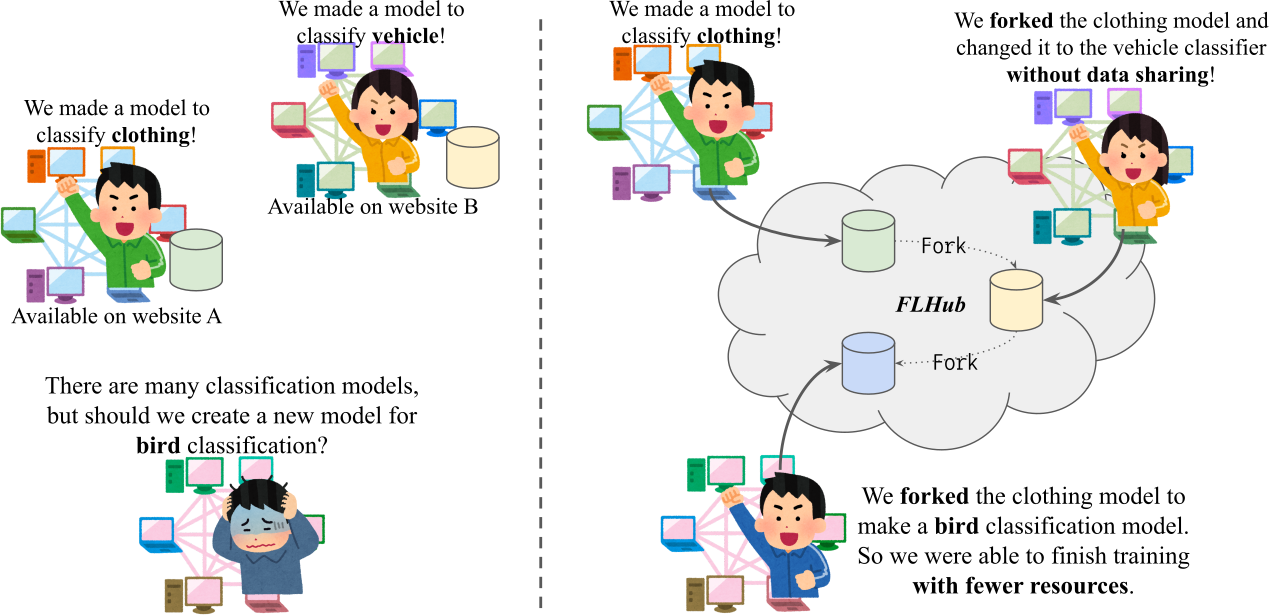}
    \caption{Sharing FL models with or without FLHub.}
    \label{fig:illustration}
\end{figure}

In federated learning, users build up the learning models with their data.
When other federated-learning model developers are to devise a training model for different applications,
they have to individually search and find the publicly available models that can be useful for their applications, as shown in Fig. \ref{fig:illustration}.
As the online model sharing service is unavailable, it is challenging to share federated learning models by contributing the training data.

We present an online federated learning model sharing service, FLHub, that provides FL model repositories where users archive, exchange, contribute and distribute FL models to solve these problems.
A federated learning model developer commits and pushes the updated model to the FLHub, where a federated learning model is archived and open to the public.
On the FLHub, a user downloads a public FL model and uploads the revised model that can be merged into the downloaded model like GitHub's pull request and merge methods.
After downloading a model, FL model users train the model with their private dataset. 

We implement the FLHub\footnote{\url{https://github.com/munhyunsu/tff-app}} service based on \texttt{TensorFlow}, the communication protocol of \texttt{gRPC} and federated learning image classification models.
The FLHub service provides APIs for online access to FL models.
In the experiment, we trained the \texttt{Fashion MNIST}, \texttt{CIFAR10}, and \texttt{Caltech Bird 200} image classification models through FLHub.
We demonstrate that the model that forked the previously trained model can finish training faster than the existing model.
In addition, the forked model takes less learning time for each federated round.
With these results, we support model publishing, asynchronous learning participation, and model fork and sharing and contribute to learning accuracy and speed among federated learning developers through FLHub.

\section{Design of FLHub (Federated Learning Hub)}

\begin{figure}[htb]
	\centering
    \includegraphics[width=1.0\columnwidth]{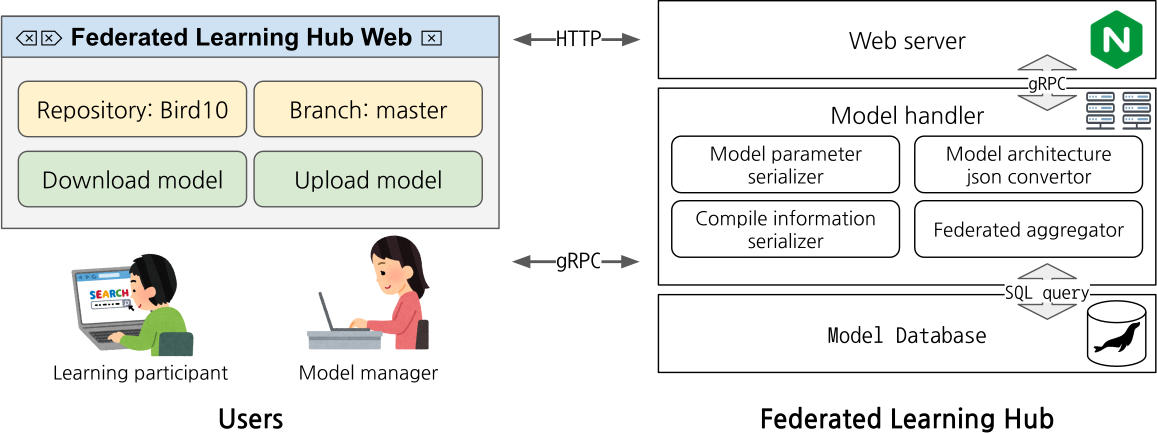}
    \caption{The architecture of FLHub}
    \label{fig:architecture}
\end{figure}

The FLHub service consists of a web server and database with the communication of \texttt{gRPC} (Figure \ref{fig:architecture}).
FLHub users are consist of model managers and learning participants with different roles like GitHub.
The model manager is responsible for maintaining and merging models, and participants upload the updated learning model with their private data.
When participants are to contribute the model by sending a pull request like GitHub, the model manager examines the learning results contributed by each participant and issues a merge command if it is effective.

The following is a simple usage scenario of the FLHub service in four steps (Fig. \ref{fig:architecture}).
\begin{enumerate}
    \item The model manager uploads and opens a new model with attributes of name, shape, initial weight, and learning method (compile information, optimizer).
    \item A private image classification dataset participant downloads the publicly available model from the FLHub server and trains the model with their data.
    \item After the participants complete the learning process asynchronously, they send the results to the FLHub server.
    \item The model manager issues a model merge command with the model name and reference version as factors if the learning result contributes to the model.
\end{enumerate}
After merging the model contributions, the HLHub web server updates the learning results to the model version, performs a federated aggregation, and records the model version in the database.

In the FLHub service, the model name and model version are essential elements to provide multiple models simultaneously and to merge the asynchronously returned training results.
The model name identifies the target model.
The model version increments each time FL model merging proceeds, and it distinguishes different model versions.
In addition, participants can prevent duplicate contributions with the model name and version.
The learning manager handles the learning results asynchronously with the reference version.
When FLHub reports multiple training results with different versions, the learning manager ignores the old learning result and selects the recent model version.
Alternatively, the model manager can create a new branch based on the old version.

\begin{figure*}[!htb]
    \centering
    \includegraphics[width=0.70\textwidth]{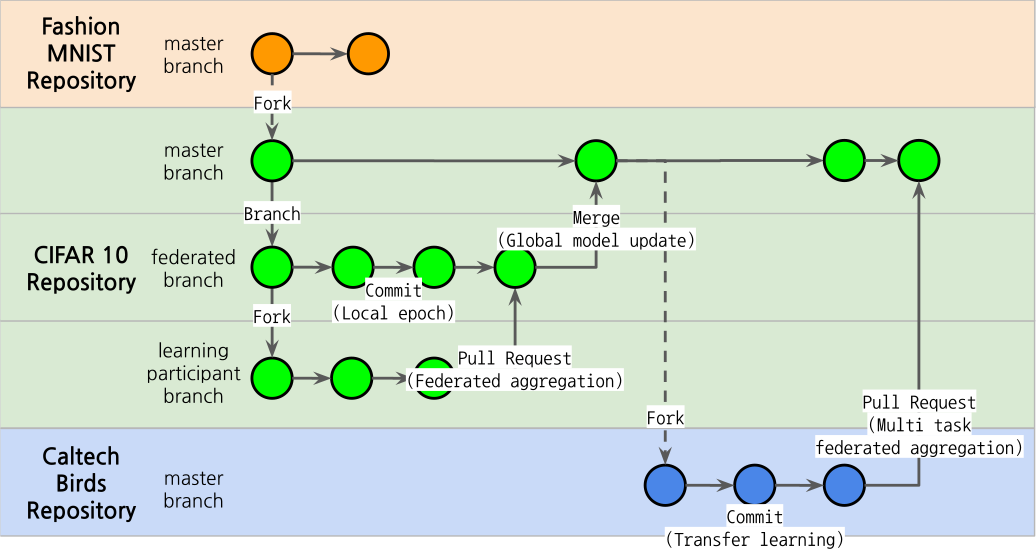}
    \caption{FLHub message flow like version control with Git.}
    \label{fig:messageflow}
\end{figure*}

FLHub identifies models and contributions by model name and version (major, minor, micro).
As shown in Fig. \ref{fig:messageflow}, FLHub identifies a model as a version and draws a network to manage asynchronous participation and model differentiation.
Therefore, when the model manager creates a branch for the next federated round (federated branch in Fig. \ref{fig:messageflow}), each participant asynchronously commits the learning result with their data and sends a pull request to the model manager (learning participant branch in Fig. \ref{fig:messageflow}).
The learning model history is displayed as a commit within the branch through local epoch, global update, etc.

\begin{table*}[htb]
\centering
\caption{FLHub command lists.}
\label{tab:command}
\small
\begin{tabular}{lll}
\hline \hline
\multicolumn{1}{c}{Command} & \multicolumn{1}{c}{Action} & \multicolumn{1}{c}{Meaning} \\
\hline
Branch & Create new branch & Ready for next federated round \\
Fork (All) & Clone the model & Prepare to participate in learning \\
Fork (Feature layer only) & Clone feature layer only & Prepare transfer learning model \\
Commit & Update model weights & Local epoch in local training \\
Pull Request (All) & Federated aggregation & Request to participate in learning \\
Pull Request (Feature layer only) & Multi task federated learning & Request to participate in learning \\
Merge & Update global model & Deploy new model with aggregated weights \\
\hline \hline
\end{tabular}
\end{table*}

\begin{lstlisting}[
caption=Sample code with FLHub,
label={lst:samplecode},
breaklines=true,]
request = federated_pb2.ModelRequest(
            NAME, VERSION)
response = stub.GetModel(request)
model = tf.keras.models.model_from_json(
          response.architecture)
model.set_weights(
        pickle.loads(response.parameter))
model.compile(
        **eval(pickle.load(response.compile)))
model.fit(datasets)
request = federated_pb2.PushModelMessage(model)
response = stub.PushModel(request)
\end{lstlisting}

Due to FLHub, it is possible to participate in asynchronous model learning by simply managing the model as a network and to create a new model by forking the public model to use in a new domain.
For example, federated learning developers and administrators can fork a learning model published in the \texttt{Fashion MNIST} repository and change it to a \texttt{CIFAR10} model, as shown in Fig. \ref{fig:messageflow}.
Since FLHub processed a series of fork-commits like transfer learning, the forked model uses the knowledge (weight parameters) learned from the source domain in the target domain.
In addition, if necessary, model managers can learn the feature layer together and perform federated multi-task learning (FMTL) with each prediction layer.
Table \ref{tab:command} shows the representative commands and operations of FLHub, and designed similarly to those of \texttt{Git}, \texttt{GitHub}, and \texttt{DockerHub} for user convenience.

FLHub is served through a web page or by directly communicating with the model handler and \texttt{gRPC} at the code level.
When a user communicates with \texttt{gRPC}, FLHub performs authentication using an API key, and an authenticated user can issue merge and pull request commands only to authorized models.
Model parameters, architecture, and compile information are serialized and stored in the RDBMS.
The model structure and weight information are described in \texttt{JavaScript object notation (JSON)} and \texttt{Binary large object (BLOB)} format.
Listing \ref{lst:samplecode} is a code sample for training a \texttt{TensorFlow} model by communicating with FLHub directly with \texttt{gRPC} in \texttt{Python 3}, which is very similar to the code for creating and training a model in \texttt{TensorFlow}.

\begin{table}[htb]
\centering
\caption{Federated Learning Hub gRPC Messages}
\small
\begin{tabular}{ll}
\hline \hline
\multicolumn{1}{c}{gRPC Messages} & \multicolumn{1}{c}{Function} \\
\hline
GetInformation & Retrieve model name, \\
 &  version \\
GetModel & Download architecture, \\
 &  parameter, and etc. of model \\
PushTrainResult & Upload learning result \\
GetStatus & Retrieve learning result \\
 & for target model \\
PushControl & Perform federated aggregation \\
 & or mark ignoring learning result \\
\hline \hline
\end{tabular}
\label{tab:api}
\end{table}

\section{Experiment}

With FLHub, a user builds federated learning applications using models that other users have trained and published.
When creating a new federated learning model, a user can complete training faster (less federated rounds) by forking the existing model in FLHub.
The federated learning models are trained \texttt{Fashion MNIST}, \texttt{CIFAR10}, and \texttt{Caltech Bird 200} image classification models.

\begin{table}[htb]
\centering
\caption{Dataset for experiments}
\small
\begin{tabular}{l|rrr}
\hline \hline
\multicolumn{1}{l}{} & \multicolumn{1}{c}{\begin{tabular}[c]{@{}c@{}}Fashion\\ MNIST\end{tabular}} & \multicolumn{1}{c}{CIFAR10} & \multicolumn{1}{c}{\begin{tabular}[c]{@{}c@{}}Caltech\\ Birds\end{tabular}} \\
\hline
Domain & Clothing & Vehicle, Animal & Bird \\
Train data & 60,000 & 50,000 & \multirow{2}{*}{17821} \\
Test data & 10,000 & 10,000 & \\
Classes & 10 & 10 & 200 \\
Clients & 3 & 5 & 10 \\
Samples & 6,400 & 6,400 & 3,200 \\
\hline \hline
\end{tabular}
\label{tab:dataset}
\end{table}

We conduct experiments to examine two questions on FLHub using table \ref{tab:dataset}.
First, does the forked model complete the training faster than the model trained in another domain compared to a randomly initialized model?
Second, between the model trained on simple data and the model trained on various data, which model should be forked and perform better?
We use \texttt{Fashion MNIST}, \texttt{CIFAR10}, and \texttt{Caltech Bird 200} data available in the \texttt{TensorFlow Datasets}.
Since a client participating in learning in the real world does not always have the same data, the client randomly sampled many data from the training dataset every round.
FLHub and experimental source code are available on Github.

\begin{figure}[htb]
	\centering
    \includegraphics[width=0.75\columnwidth]{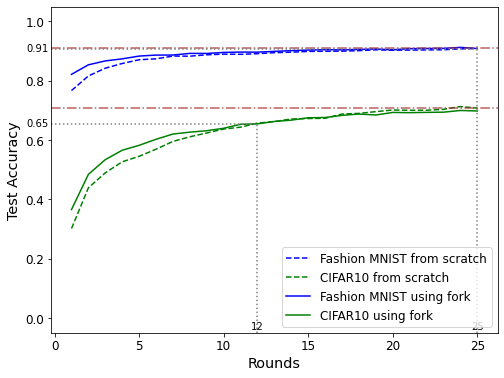}
    \caption{Comparison of test accuracy comparison between FL model forked in FLHub and FL model built from scratch.}
    \label{fig:accuracy}
\end{figure}

We fork a model already trained in another domain and modify it into the target application model.
Figure \ref{fig:accuracy} shows the accuracy according to the rounds of the new model from scratch and the forked model when building the \texttt{Fashion MNIST} classifier and the \texttt{CIFAR10} classifier.
We update the forked \texttt{Fashion MNIST} classifier with the \texttt{CIFAR10} decimal classifier after 50 federated rounds.
We also generate the forked \texttt{CIFAR10} classifier by forking the \texttt{Fashion MNIST} classifier after 50 federated rounds.
Both the forked \texttt{Fashion MNIST} model and the forked \texttt{CIFAR10} model achieved the same accuracy in fewer rounds than the original model.

\begin{figure}[htb]
	\centering
    \includegraphics[width=0.75\columnwidth]{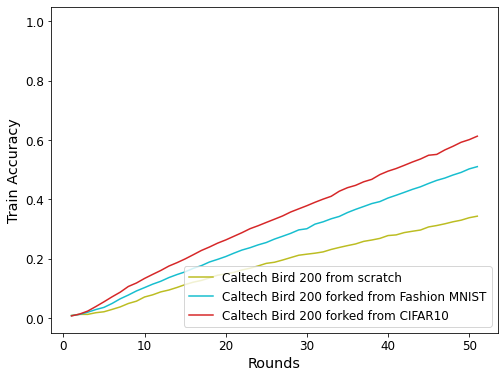}
    \caption{Comparison of training accuracy of original model, a forked Fashion MNIST in FLHub, and a forked CIFAR 10 in FLHub.}
    \label{fig:fork}
\end{figure}

We fork the \texttt{Fashion MNIST} model and the \texttt{CIFAR10} model trained for the same purpose, respectively, to examine the importance of the target of the forking model.
FLHub converts a model initially a 10 class classifier into a 200 class classifier by changing the last prediction layer.
This task can be performed more precisely at the code level or simply on the web.
We trained \texttt{Caltech Bird 200} data on \texttt{Caltech Bird} from scratch, \texttt{Caltech Bird} fork from \texttt{Fashion MNIST}, and \texttt{Caltech Bird} fork from \texttt{CIFAR10} models, and looked at models that train better with train accuracy.

\texttt{Caltech Bird} fork from \texttt{CIFAR10}, which forked a complex model, had enormous learning per round (Figure \ref{fig:fork}).
\texttt{CIFAR10} is a dataset with various vehicles and animals, and since it is more complex than \texttt{Fashion MNIST}, a model requires more feature extraction for classification.
This pre-learned feature extraction knowledge (model parameter) is forked like transfer learning so that model can find the optimal solution faster for each round.
Conversely, if there is no such knowledge (scratch) or minor (\texttt{Fashion MNIST}), the optimal solution is found relatively slowly.

\section{Conclusion}

This paper presents an online federated learning model sharing service, FLHub, that provides FL model repositories where users archive, exchange, contribute and distribute FL models.
Through the FLHub API, FL model developers and users can share FL models and learn them asynchronously.
In addition, the user can fork the previously trained model and use it in other domains.
We demonstrated that the forked model is more accurate and learns faster than the newly created model.
We contributed model publishing, asynchronous learning participation, model fork and sharing, and learning accuracy and speed through FLHub.

\section{Acknowledgments}
This work was supported by Institute for Information \& communications Technology Planning \& Evaluation (IITP) grant funded by the Korea government (MSIT) (No.2019-0-01343, Training Key Talents in Industrial Convergence Security)

\bibliography{main}

\end{document}